\documentclass[10pt]{article}

\usepackage{longtable}

\usepackage{tabularx}

\usepackage{cancel}

\usepackage{lettrine}

\usepackage[margin=1in]{geometry} 
\usepackage[utf8]{inputenc}
\usepackage[T1]{fontenc}

\usepackage[english]{babel}

\usepackage{dashrule} 

\usepackage{extarrows}

\usepackage{amsmath,amsthm,amscd,amssymb,amsfonts,bm}

\usepackage{bbm, mathrsfs, upgreek} 

\usepackage{xcolor}

\usepackage{sectsty}
\sectionfont{\normalfont}

\usepackage{stackengine}

\usepackage{tikz-cd, tikz, pgfplots,pgf}
\pgfplotsset{compat=1.15}
\usetikzlibrary{arrows}

\usepackage{caption}
\usepackage{wrapfig,graphicx}
\graphicspath{ {./figures/} } 
\usepackage[section]{placeins}

\numberwithin{equation}{section} 
\numberwithin{figure}{section} 

\stackMath

\usepackage{enumitem}

\usepackage{mathpazo}

\usepackage{setspace}
\usepackage{footmisc}
\usepackage{hyperref}
\hypersetup{
    colorlinks=true,
    linkcolor=blue,
    filecolor=magenta,      
    urlcolor=cyan,
    citecolor=blue,
}
\usepackage{doi}
\usepackage[numbers, sort&compress]{natbib}
 \bibpunct[, ]{[}{]}{,}{n}{}{,}%
 \def\bibfont{\small}%

\allowdisplaybreaks

\counterwithin{table}{section}

\usepackage[most]{tcolorbox}
\definecolor{customblue}{rgb}{0.6, 0.8, 1.0}
\newtcolorbox{promptbox}[1][]{
  colback=customblue!40,
  colframe=customblue!80!black,
  sharp corners,
  boxrule=0.5mm,
  shadow={1mm}{-1mm}{0mm}{black!50!white},
  fonttitle=\bfseries,
  title=#1,
  breakable,
  enhanced,
  borderline={0.5mm}{0mm}{customblue!80!black},
  rounded corners
}

\newtcolorbox{problembox}[1][]{
  colback=red!40,
  colframe=red!80!black,
  sharp corners,
  boxrule=0.5mm,
  shadow={1mm}{-1mm}{0mm}{black!50!white},
  fonttitle=\bfseries,
  title=#1,
  breakable,
  enhanced,
  borderline={0.5mm}{0mm}{red!80!white},
  rounded corners
}

\begin{document}
{\linespread{1}\selectfont
\title{\linespread{1}\selectfont Dual Traits in Probabilistic Reasoning of Large Language Models}
\author{\linespread{1}\selectfont Shenxiong Li\footnote{University of Rochester, Simon Business School, sli73@ur.rochester.edu}, Huaxia Rui\footnote{University of Rochester, Simon Business School, huaxia.rui@simon.rochester.edu}}
\date{}
\maketitle}

\renewcommand{\thetable}{\arabic{table}}
\setcounter{table}{0}

\begin{abstract}
We conducted three experiments to investigate how large language models (LLMs) evaluate posterior probabilities. Our results reveal the coexistence of two modes in posterior judgment among state-of-the-art models: a normative mode, which adheres to Bayes' rule, and a representative-based mode, which relies on similarity --- paralleling human System 1 and System 2 thinking. Additionally, we observed that LLMs struggle to recall base rate information from their memory, and developing prompt engineering strategies to mitigate representative-based judgment may be challenging. We further conjecture that the dual modes of judgment may be a result of the contrastive loss function employed in reinforcement learning from human feedback. Our findings underscore the potential direction for reducing cognitive biases in LLMs and the necessity for cautious deployment of LLMs in critical areas.
\end{abstract}

The remarkable advancements in large language models (LLMs) have ushered in a new era where these models rival human expertise across domains like academia, law, medicine, and finance \citep{doi:10.1073/pnas.2404328121, cui2024chatlawmultiagentcollaborativelegal, wang2021cloud,li2023natural, wu2023bloomberggptlargelanguagemodel, yang2023fingptopensourcefinanciallarge}. Central to decision-making in these fields is assessing the posterior probability $P(H|E)$ of a hypothesis $H$ that a specific object belongs to a predefined class, given the observed evidence $E$ of the object. In this study, we explore how LLMs judge this posterior probability.

We define a \textit{posterior function} $f(H, E) \in [0,1]$, which maps the hypothesis $H$ and evidence $E$ to the assessed posterior probability $P(H|E)$. Drawing from psychology literature \citep{KAHNEMAN1972430, doi:10.1126/science.185.4157.1124}, we consider two modes of judgment: \textit{normative} ($f_{\text{norm}}$), aligning with Bayes' rule, and \textit{representativeness} ($f_{\text{rep}}$), which evaluates how similar the evidence $E$ is to a typical or prototypical\footnote{Representativeness can be constructed through typicality or prototypicality. Typicality describes the common or average case of the class, whereas prototypicality embodies the most idealized and iconic version of the class. For instance, a typical example of a physicist is a smart man who likes math and physics, while a prototypical example of a physicist is Stephen Hawking.} example of the class $H$. A higher similarity corresponds to a higher assessed posterior probability. This study comprises three experiments with progressively stricter conditions, reducing the information available for posterior likelihood assessment. The \textit{structured test} provides all information needed for normative judgment, the \textit{semi-structured test} omits the diagnosticity of evidence, and the \textit{unstructured test} requires LLMs to recall all components of Bayes' rule. Results reveal that LLMs' judgments shift from $f_{\text{norm}}$ to $f_{\text{rep}}$, reflecting \textit{dual traits} analogous to human System 1 and System 2 thinking \citep{Stanovich2000}.

Previous studies on LLMs show that while they sometimes align with normative principles, they are also prone to cognitive biases \citep{echterhoff2024cognitivebiasdecisionmakingllms, doi:10.1073/pnas.2218523120, hagendorff2023human}. This study moves beyond bias detection to investigate the basis upon which LLMs assess probabilities. In addition to identifying deviations from $f_{\text{norm}}$, we demonstrate the presence of another tangible judgment mode $f_{\text{rep}}$. Our findings shed new light on the debate regarding whether LLMs operate as System 1 (intuitive) or System 2 (analytical) thinkers, revealing a coexistence of both modes within LLMs. Moreover, through three experiments, we investigate under what conditions LLMs demonstrate $f_{\text{norm}}$ and $f_{\text{rep}}$, as well as the point at which the transition occurs. This has important practical implications for the integration of LLMs into various critical fields.

Importantly, our results suggest that prior evaluations focusing solely on the accuracy of LLMs' responses may mischaracterize deviations from normative reasoning. Low accuracy can arise from computational errors, limited background knowledge, or inherent randomness, rather than from deviations from normative principles. Previous studies have at best documented that LLMs may be susceptible to probabilistic reasoning errors. In contrast, our experiments provide more structured insights. First, we demonstrate that LLMs' posterior judgments are insensitive to changes in base rates, clearly indicating that their reasoning do not follow Bayes' rule. Second, we find a nearly linear positive correlation between posterior judgments and similarity judgments, suggesting that these judgments are based on representativeness. Nonetheless, the susceptibility to probabilistic reasoning errors poses challenges in interpreting the results. However, in a structured testing environment, we show that state-of-the-art LLMs exhibit significant potential in \(f_{\text{norm}}\) by demonstrating strong signs of overcoming previously documented errors.

It's intriguing to consider how LLMs develop dual traits in probabilistic reasoning. As we conclude this article, we reflect on this and propose exploring the relationship between cognitive biases in LLMs and the contrastive loss function employed in reinforcement learning from human feedback. We believe that understanding this connection could provide a pathway to mitigating cognitive biases in LLMs, thereby enhancing the reliability of their reasoning and responses.

\subsection*{Structured Test}
The structured test adapts and modifies problems from prior studies \citep{KAHNEMAN1972430, tversky1982causal, BARHILLEL1980211}, detailed in (Supplementary Information Section 2.1). Fifteen models were tested, categorized as \textit{state-of-the-art} (e.g., GPT-4, Gemini-1.5-Pro), \textit{affordable} (e.g., Claude-3-Sonnet, Open-Mixtral-8x22B), and \textit{small or outdated} (e.g., GPT-3.5, Mistral-Small). Results (Extended Data Table \ref{table:reasoning_main}) show that state-of-the-art models excel in probabilistic reasoning, outperforming less advanced models and humans. We also found that less advanced models often made computational errors, misunderstood questions, or failed due to insufficient knowledge. Therefore, subsequent experiments focus on state-of-the-art models.

\subsection*{Semi-Structured Test}
The semi-structured test provides only the base rate, restricting the evaluation of LLMs' likelihood judgments to the posterior's deviation from the prior. However, a significant deviation is normatively valid if the evidence's diagnosticity, $D(H, E)$, differs substantially from one. Denoting the normative posterior as $f_{\text{norm}}(B, H, E)$ with base rate $B$, hypothesis $H$, and evidence $E$, the posterior odds can be expressed as:

{\linespread{0.9}\selectfont
$$O(B, H, E) \equiv \frac{f_{\text{norm}}(B, H, E)}{1 - f_{\text{norm}}(B, H, E)} = \frac{B}{1 - B} \times D(H, E).$$}

\vspace{-3pt}

\noindent Posterior odds regress to prior odds $\frac{B}{1-B}$ only when $D(H, E)$ approaches one. Thus, isolating evidence diagnosticity is crucial, which can be achieved using the following prompt structure.

\begin{promptbox}[\small{Prompt Structure}]
\small \linespread{0.5}\selectfont
A study aimed to identify common personality traits among successful academics was conducted. In this study, a panel of professional psychologists interviewed and administered personality tests to 75 \textit{(25)} tenured and successful professors in computer science and 25 \textit{(75)} tenured and successful professors in the humanities. Based on these personality tests, descriptions were written for each of the 100 professors.
\vspace{5pt}

You will be presented with one such description, chosen at random from the 100 available descriptions.
\vspace{5pt}

Here is the description: \textit{[A description of Jason]}
\vspace{5pt}

Now, please indicate your probability that Jason is one of the computer science professors in the sample.
\end{promptbox}

The prompt structure includes two base rate settings. In the high base rate setting with $75\%$ computer science professors, $B_{h} = 0.75$, yielding prior odds $\frac{B_{h}}{1-B_{h}} = 3$. In the low base rate setting, $B_{\ell} = 0.25$, with prior odds $\frac{B_{\ell}}{1-B_{\ell}} = \frac{1}{3}$. Since the diagnosticity of evidence $D(H, E)$ depends only on the evidence and hypothesis, it remains unaffected by the base rate. Thus, for normative likelihood judgments:

{\linespread{0.8}\selectfont
\begin{itemize}
\itemsep0em
\item High base rate: $O(B_{h}, H, E) = 3D(H, E)$.
\item Low base rate: $O(B_{\ell}, H, E) = \frac{1}{3}D(H, E)$.
\end{itemize}}

\vspace{-5pt}

The ratio of posterior odds simplifies to:

{\linespread{0.9}\selectfont
$$\frac{O(B_{h}, H, E)}{O(B_{\ell}, H, E)} =\frac{3D(H,E)}{\frac{1}{3}D(H,E)}=9.$$}

\vspace{-3pt}

\noindent This ratio remains constant at $9$ for any fixed description, assuming normative evaluation of likelihoods.

Table \ref{table:json_main} summarizes the experimental results\footnote{Detailed descriptions and prompts are in Supplementary Information Section 2.2.}. For the uncharacteristic description, 4 of 6 LLMs display normative performance with a posterior odds ratio near $9$. Although Gemini-1.5-Pro and Mistral-Large deviate slightly, their posterior probabilities still respond to base rate changes. Conversely, for representative descriptions, the posterior odds ratios for all LLMs approximate $1$ instead of $9$, indicating non-normative judgment. Moreover, LLMs consistently assign higher posterior probabilities for computer science descriptions and lower for humanities, suggesting representativeness-based likelihood judgments.

{\tiny
\linespread{0.5}\selectfont
\begin{table}[htbp]
\centering
\caption{Posterior Probabilities in the Jason Test}
\label{table:json_main}
\resizebox{\textwidth}{!}{
\begin{tabular}{lccccccc}
\hline
\hline
\multicolumn{1}{c}{} &
  \textbf{gpt-4o} &
  \textbf{gpt-4} &
  \textbf{\begin{tabular}[c]{@{}c@{}}claude-\\ 3-opus\end{tabular}} &
  \textbf{\begin{tabular}[c]{@{}c@{}}gemini-\\ 1.5-pro\end{tabular}} &
  \textbf{\begin{tabular}[c]{@{}c@{}}mistral-\\ large\end{tabular}} &
  \textbf{\begin{tabular}[c]{@{}c@{}}llama-3-70b-\\ instruct\end{tabular}} &
  \textbf{\begin{tabular}[c]{@{}c@{}}Human \\ Average\end{tabular}} \\ \hline
\multicolumn{8}{c}{\textbf{Uncharacteristic Description}} \\ \hline
High base rate (75\%) posterior & 0.75 & 0.75 & 0.75 & 0.71 & 0.75 & 0.75 & 0.63 \\
Low base rate (25\%) posterior  & 0.25 & 0.25 & 0.25 & 0.27 & 0.5  & 0.25 & 0.43 \\
Difference in posteriors            & 0.5  & 0.5  & 0.5  & 0.43 & 0.25 & 0.5  & 0.20 \\
Ratio of posterior odds         & 9    & 9    & 8.86 & 6.36 & 3    & 9    & 2.25 \\ \hline
\multicolumn{8}{c}{\textbf{Representative Description --- Computer Science}} \\ \hline
High base rate (75\%) posterior & 0.78 & 0.82 & 0.76 & 0.89 & 0.84 & 0.88 & 0.81 \\
Low base rate (25\%) posterior  & 0.74 & 0.79 & 0.65 & 0.76 & 0.77 & 0.83 & 0.67 \\
Difference in posteriors              & 0.04 & 0.03 & 0.11 & 0.13 & 0.07 & 0.05 & 0.13 \\
Ratio of posterior odds         & 1.25 & 1.18 & 1.68 & 2.58 & 1.62 & 1.52 & 2.02 \\ \hline
\multicolumn{8}{c}{\textbf{Representative Description --- Humanity}} \\ \hline
High base rate (75\%) posterior & 0.24 & 0.21 & 0.21 & 0.24 & 0.3  & 0.2  & 0.32 \\
Low base rate (25\%) posterior  & 0.24 & 0.2  & 0.17 & 0.2  & 0.3  & 0.13 & 0.19 \\
Difference in posteriors              & 0.01 & 0    & 0.04 & 0.04 & 0    & 0.07 & 0.13 \\
Ratio of posterior odds         & 1.04 & 1.02 & 1.29 & 1.28 & 1    & 1.6  & 1.99 \\ \hline
\hline
\multicolumn{8}{p{1\textwidth}}{Note: the human average column was computed using the results interpreted from \citep[p.242, Figure 1]{Kahneman1973} and a series of results from replications of the Lawyer-Engineer problems summarized in \citep[p.3, Table 1]{Koehler1996}. For the human results, the base rate was manipulated to be (0.7, 0.3), thereby indicating that the normative ratio should be around 5.44.}
\end{tabular}}
\end{table}
\FloatBarrier}

\subsection*{Unstructured Test}
The unstructured test includes three graduate study fields—agricultural and veterinary science (field A), business administration (field B), and computer science (field C)—along with a description of a person named Adam. Field selection and descriptions were designed to exploit prior discrepancies. For instance, $P(B)$ consistently exceeds $P(A)$ by 4 to 10 times, and $P(C)$ is 4 to 7 times greater than $P(A)$. To test representativeness-based judgments, we crafted a description closely aligned with field A, the field with the smallest prior. If likelihood judgments rely on representativeness, field A should have the highest posterior. Conversely, normative reasoning, which integrates priors and evidence diagnosticity, would result in less extreme posterior judgments.

LLMs were tasked with estimating Adam's posterior likelihood of belonging to each of the three fields and rating his similarity to a graduate student in each field. Pearson and Spearman correlations were computed between base rates, posterior probabilities, and similarity scores. Table \ref{table:corr1} shows a significant negative correlation between priors and posteriors, and a significant positive correlation between posteriors and similarity scores. This divergence can occur normatively only when the diagnosticity $D$ deviates largely from $1$. To assess the diagnosticity, LLMs were tasked with judging the inverse probabilities $P(E|H)$ and $P(E|\neg H)$. Results (Extended Data Table \ref{table:diagno}) indicate diagnosticity values near $1$, suggesting the description is non-diagnostic. Thus, normative posteriors should align with priors, but LLMs' judgments deviate, closely mirroring similarity scores.

{\footnotesize
\linespread{0.5}\selectfont
\begin{table}[htbp]
    \centering
    \caption{Correlations between Prior, Posterior and Similarity}
    \label{table:corr1}
    \begin{tabular}{l|cc|cc}
        \hline
        \hline
         & \multicolumn{2}{c|}{\textbf{Prior vs. Posterior}} & \multicolumn{2}{c}{\textbf{Posterior vs. Similarity}} \\ 
        \hline
        Model Name     & Spearman & Pearson  & Spearman & Pearson \\ 
        \hline
        gpt-4o         & -0.88*** & -0.95*** & 0.91***  & 0.90*** \\
        gpt-4-turbo    & -0.88*** & -0.88**  & 0.90***  & 0.95*** \\
        claude-3-opus  & -0.90*** & -0.92*** & 0.87***  & 0.92*** \\
        gemini-1.5-pro & -0.84**  & -0.94*** & 0.90***  & 0.96*** \\
        mistral-large  & -0.87*** & -0.95*** & 0.94***  & 0.98*** \\ 
        \hline
        \hline
        \multicolumn{5}{p{0.6\textwidth}}{{\footnotesize Note: two-sided Fisher's p-value, significance levels: ***$p < 0.001$, **$p < 0.01$, *$p < 0.05$.}}
    \end{tabular}
\end{table}
\FloatBarrier}

Correlations between inverse probabilities, posteriors, and similarities (Extended Data Table \ref{table:corr2}) reveal strong positive correlations, with judged inverse probability and posteriors both nearly perfectly correlated with similarity.

\subsection*{Prompt Engineering}
In the semi-structured test, altering a single word in the prompt could reverse biased behavior by compelling LLMs to compute step-by-step using Bayes' rule (Extended Data Table \ref{tab:semi_modify}). Key changes included:

{\linespread{0.5}\selectfont \begin{itemize} \itemsep0em \item \textit{indicate} $\rightarrow$ \textit{compute} \item \textit{probability} $\rightarrow$ \textit{posterior probability} \item \textit{probability} $\rightarrow$ \textit{posterior} 
\end{itemize}}

\vspace{-5pt}

\noindent This, combined with structured test results, highlights the importance of explicitly guiding LLMs to perform multi-step Bayes' rule computations.

Guiding LLMs to perform Bayes' rule, however, is not always straightforward. In the unstructured test, we implemented four iterative prompt modifications (Supplementary Information Section 2.4), with each subsequent prompt increasingly emphasizing the application of Bayes' rule. Despite this, GPT-4o still relied on evaluating Adam's resemblance to graduate students, merely acknowledging Bayes' rule without fully applying it. The observed correlations do not vary in magnitude or direction (Extended Data Table \ref{tab:un_modify}). Only in the final modification did it compute the posterior probability step-by-step, but even then, inconsistencies arose. For example, when separately asked, GPT-4o provided priors for field A between $2\%$ and $4\%$, yet during posterior judgment, it recalled priors fluctuating between $10\%$ and $30\%$. Similar discrepancies appeared for inverse probabilities $P(E|\neg H)$, but the inaccurate recall of base rates posed a more severe challenge.

\subsection*{Discussion}

\subsubsection*{Practical Implications}
Representativeness judgment has historically led to human errors in diagnoses and decisions. As LLMs are increasingly integrated into critical fields, it is vital to understand how they judge posterior probabilities. This study highlights the coexistence of $f_{\text{norm}}$ and $f_{\text{rep}}$ in LLMs, cautioning against uncritical reliance on their outputs. Users of LLM-integrated tools must recognize that LLM likelihood judgments, such as diagnosing illnesses, can be biased and should account for factors like base rates (e.g., the rarity of diseases).

To improve domain-specific LLMs, training data should include distributional information relevant to the domain. For instance, a medical diagnosis LLM should be fine-tuned to account for disease prevalence, while a financial analysis LLM should prioritize objective metrics such as returns and market rates over subjective features like corporate culture or CEO characteristics. Additionally, our experiments underscore the importance of including base rates in prompts, as LLMs often struggle to recall them. Finally, it is worth noting that, while prompt engineering is critical, identifying patterns that reliably trigger LLMs to apply $f_{\text{norm}}$ can be challenging and may vary in complexity.

\subsubsection*{A Conjecture on Bias Origins}
While our primary research question has been addressed, two key unresolved questions remain:

{\linespread{0.8}\selectfont
\begin{enumerate}
\itemsep0em
\item What drives the coexistence of the two judgment systems in LLMs?
\item Why does the structured test uniquely prompt LLMs to automatically initiate normative computation via Bayes' rule?
\end{enumerate}}

\vspace{-10pt}

\noindent We posit that the answer to these two questions is intricately linked to the supervised fine-tuning (SFT) of LLMs. The following discussion primarily revolves around the well-documented Llama 2 and Llama 3 families \citep{touvron2023llama2openfoundation, dubey2024llama3herdmodels}. Notably, the core principles of SFT in the Llama series were first introduced by the OpenAI team during their training of a summarizer using GPT-3 \citep{NEURIPS2020_1f89885d} and their development of GPT3-Instruct \citep{NEURIPS2022_b1efde53}. Thus, our discussion also provides insights into the workings of closed-source models such as GPT-4 and GPT-4o.

One of the most critical steps in the SFT phase is the training of a reward model. This model is the copy of the LLM following pre-training and possible initial instruction tuning\footnote{Instruction tuning is performed on a dataset where each data point includes the context $x$ and a ground-truth output $y$. The context is augmented with human-written instructions, such as ``please provide your best estimates of the probability...''. The loss function utilized is the standard autoregressive cross-entropy loss, with back-propagation applied only to the output.}. Here, the classification head used for next-token prediction is replaced with a regression head designed to output a scalar reward. The training process leverages the following binary ranking loss function:

{\linespread{1}\selectfont
$$\mathcal{L} \equiv -\log(\sigma(r_{\theta}(x, y_{c}) - r_{\theta}(x, y_{r}))),$$}

\vspace{-3pt}

\noindent where $\sigma$ represents the sigmoid function, $r_{\theta}(x,y)$ denotes the scalar score output based on the context prompt $x$ and completion $y$ with model weights $\theta$. The term $y_{c}$ stands for the preferred response favored by human annotators, while $y_{r}$ refers to the rejected counterpart.

Two critical observations can be made about this loss function. First, it essentially disregards the base rate, because it only considers one instance per class. Second, this loss function aligns with the loss employed in the framework of contrastive learning, where $x$ is an anchor, $y_{c}$ is a positive sample, and $y_{r}$ is a negative sample \citep{arora2019constrast, NEURIPS2020_d89a66c7}. Hence, it inherently drives the model to identify and amplify the features that make $y_{c}$ preferable to $y_{r}$. This selective emphasis enables the model to adeptly recognize the representative cues that align with desired outcomes, but it also leads the model to become overly attuned to specific features distinguishing these pairs, potentially overfitting to the representativeness. Subsequently, the reward model's overfitting is transferred to the LLM through reinforcement learning by proximal policy optimization (PPO) \citep{schulman2017proximalpolicyoptimizationalgorithms}. It has been showed that the PPO algorithm is equivalent to directly training the LLM using a loss function whose ``gradient increases the likelihood of the preferred completion and decreases the likelihood of dispreferred completions'' \citep[p.5]{NEURIPS2023_a85b405e}\footnote{The cited paper developed direct preference optimization (DPO). It directly utilizes the LLM as a reward model with a loss function whose gradient performs equivalently to contrastive learning. The paper demonstrates that DPO and PPO are mathematically equivalent up to a minor reparametrization, but DPO exhibits greater stability in terms of optimization. This algorithm was subsequently applied in the training of the Llama-3 family. The Meta team found that ``DPO required less compute for large-scale models and performed better'' \citep[p.16]{dubey2024llama3herdmodels}.}, an idea consistent with contrastive learning.

Does the training dataset indeed contain sufficient instances for such overfitting to occur? The answer is affirmative. A considerable number of tasks within the SFT dataset involve classification, extraction, and summarization \citep{zhao2023survey, zhang2024instruction, NEURIPS2022_b1efde53}. These tasks essentially involve identifying and leveraging distinct features of texts. Repeated training on pairs of preferred and rejected results for these tasks inevitably leads the model to overly favor the representativeness of evidence. Notably, the loss function entirely neglects the base rate, thereby establishing the system of $f_{\text{rep}}$. 

On the other hand, another considerable portion of the SFT data consists of mathematical problems with detailed solutions. By fine-tuning on a vast corpus of probability and statistics exercises, LLMs learn to recognize patterns in these problems and their solutions. This enables LLMs to identify similar problems and autonomously solve them by following memorized steps and simply plugging in the numbers. This explains their exceptional and autonomous performance in the structured test. In general, we posit that it is the memorization of a substantial collection of mathematical problems that endows LLMs with the system of $f_{\text{norm}}$.

Finally, we note that the tuning on mathematical problems only taught LLMs the statistical regularities between the structure of a problem and its corresponding solution. Hence, they may struggle to apply $f_{\text{norm}}$ when the prompt lacks some key patterns, necessitating explicit human instruction for LLMs to perform multi-step computations using Bayes' rule in the semi-structured and unstructured tests. Without explicit instruction, LLMs would invoke $f_{\text{rep}}$ and concentrate primarily on the representativeness of the evidence, even if base rate is included in the prompt.

In conclusion, we address the necessity of further exploring the relationship between cognitive biases, probabilistic reasoning, and supervised fine-tuning in LLMs. Specifically, the use of a contrastive loss function can be considered as an initial step towards mitigating bias in LLMs.

\subsection*{Methods}

\subsubsection*{Experimental Designs}
The evaluation of likelihood judgments should not be based on a binary right-or-wrong metric, as they inherently represent subjective probabilities. Accuracy can only be genuinely assessed when a question includes a full set of Bayesian information, akin to structured tests --- an often impractical scenario. Therefore, to test if LLMs deviate from normative reasoning, some Bayesian information must be omitted. Evaluation should thus focus on the extent of likelihood judgments relative to similarity judgments, necessitating a manually crafted potential extremity. In the semi-structured tests, both types of inverse probability were omitted, and the descriptions were crafted to be as extreme in representativeness as possible. In the unstructured test, neither base rate nor the diagnosticity of evidence was provided. And our design focuses on the extremes between the high representativeness of the description and the low base rate of the field associated with this description. Moreover, we incorporated some noises in the description to ensure that it is representative but not diagnostic.

\subsubsection*{Inherent Randomness in LLM's Responses}
The answer provided by an LLM can be considered as a random variable $X$, so querying each question only once is problematic. Therefore, our research focuses the expected value $\mathbb{E}(X)$. To achieve this, each question was posed a predetermined number of times (referred to as the questioning round) to ensure that, with $99\%$ confidence, the sample mean closely approximates $\mathbb{E}(X)$. The methodology for determining the questioning round, as well as the specific questioning round used for each model and each question, are detailed in (Supplementary Information Section 3).

\subsubsection*{Rotational Design in Unstructured Test}
To increase the sample size for statistical analysis, the unstructured test employs a rotational design for all the questions. Each question is posed $7$ times: three times with a single field (A, B, C), three times with two-field combinations (AB, AC, BC), and once with all three fields (ABC). The significance of correlations was determined using a two-sided Fisher's p-value due to the small sample size. To mitigate finite-sample variance, the permutation distribution is generated after transforming the correlation coefficient $r$ via Fisher's $z$-transformation: $r \mapsto \frac{1}{2}\log\frac{1+r}{1-r}$. Prompts details are provided in (Supplementary Information Section 2.3).

\subsubsection*{Metalinguistic Prompts and Data Contamination}
Our methodology for evaluating LLMs' probabilistic judgment hinges on the use of metalinguistic prompts which may not effectively capture the true probability within LLMs. It is crucial to highlight that state-of-the-art LLMs are predominantly closed-source, making access to their logits impossible. Despite this, we sought to confirm the robustness of our results by prompting LLMs to complete sentences. For example, in the semi-structured test, we used the prompt ``Jason is one of the \underline{\ \ \ \ \ }'' and observed consistent behaviors.

Regarding data contamination, each question and description in the semi-structured and unstructured tests is uniquely tailored. Most questions in the structured tests have been modified in terms of context and numerical values.

\subsubsection*{Other Considerations}
All models in the experiment have undergone instruction tuning. No instruction prompts were given during this experiment. All sampling parameters are left as their default values. For all questions in the experiment, we specified a dictionary format for responses in the prompt, such as \textit{``please answer the question in the following format: \{answer: your estimate\}.''} This requirement is essential to avoid the need of spending considerable time identifying all possible parsable patterns in LLMs' answers. For questions with more than one choice, the choices presented in the prompt and in the required output dictionary were both shuffled while maintained the same order. The structured test was conducted from 08/01/2024 to 08/05/2024. The semi-structured test was administered from 07/17/2024 to 07/18/2024, and unstructured was conducted on 08/14/2024.

\subsection*{Data and Code Availability}
The experimental code and specific responses from LLMs are accessible on the associated GitHub page\footnote{\small{https://github.com/Jacobsonradical/LLM-Dual-Trait}}.

\newpage

\bibliographystyle{informs2014} 
\bibliography{ref_llm-integrated-tools, ref_human-rep-bias, ref_llm-bias, ref_general} 

\newpage

\renewcommand{\thetable}{A.\arabic{table}}
\setcounter{table}{0}

\section*{Extended Data}

\begin{table}[htbp]
\centering
\caption{Accuracy Rate of Each Question in Structured Test}
\label{table:reasoning_main}
\begin{tabular}{lcccc}
& State of the Art & Affordable & Small or Outdated & Human \\
\hline
\hline
Q1  & \textbf{0.95} & 0.66          & 0.25 & 0.18            \\
Q2  & \textbf{0.93} & 0.65          & 0.18 & N/A             \\
Q3  & \textbf{0.94} & 0.66          & 0.6  & 0.44            \\
Q4  & \textbf{1}    & 0.1          & 0.16 & 0.2             \\
Q5  & \textbf{0.97} & 0.59          & 0.51 & 0.43            \\
Q6  & 0.7          & \textbf{0.84} & 0.29 & 0.375           \\
Q7  & \textbf{0.96} & 0.67          & 0.11 & \textless 0.35  \\
Q8  & \textbf{0.75} & 0.45          & 0.15 & \textless 0.6   \\
Q9  & 0.94          & \textbf{0.97} & 0.58 & \textless 0.4   \\
Q10 & \textbf{0.39} & 0.33          & 0.06 & N/A             \\
Q11 & \textbf{1}    & 0.71          & 0.09 & \textless 0.48 \\
Q12 & \textbf{0.94} & 0.69          & 0.07 & N/A             \\
Q13 & \textbf{1}    & \textbf{1}    & 0.87 & \textless 0.5   \\
Q14 & \textbf{1}    & \textbf{1}    & 0.97 & \textless 0.5   \\
Q15 & \textbf{1}    & \textbf{1}    & 0.99 & \textless 0.5   \\
Q16 & \textbf{1}    & \textbf{1}    & 0.8  & \textless 0.5   \\
\hline
\hline
\multicolumn{5}{p{0.4\textwidth}}{{\footnotesize Note: bold numbers indicate the best performance.}}
\end{tabular}
\end{table}

\textbf{State of the Art}: GPT-4o, GPT-4, Gemini-1.5-Pro, Claude-3-Opus, Mistral-Large. 

\textbf{Affordable}: Claude-3-Sonnet, Gemini-1.5-Flash, Mistral-Medium, Open-Mixtral-8x22B. 

\textbf{Small or Outdated}: GPT-3.5, Gemini-1.0-Pro, Claude-3-Haiku, Open-Mixtral-8x7B, Open-Mistral-7B, Mistral-Small. 

\

\begin{table}[htbp]
    \centering
    \caption{Computed Diagnosticity of Evidence}
    \label{table:diagno}
    \resizebox{\textwidth}{!}{
    \begin{tabular}{lcccccccccccccc}
        \hline\hline
        \multicolumn{1}{c}{} & \multicolumn{4}{c}{D(E,A)} &  & \multicolumn{4}{c}{D(E,B)} &  & \multicolumn{4}{c}{D(E,C)} \\ 
        \cline{2-5} \cline{7-10} \cline{12-15} 
        & A & AB & AC & ABC &  & B & AB & BC & ABC &  & C & AC & BC & ABC \\ 
        \cline{2-5} \cline{7-10} \cline{12-15} 
        gpt-4o         & 1.09  & 1.04 & 1.02 & 1.01 &  & 1.37  & 1.18 & 1.55 & 1.04 &  & 1.50 & 1.28 & 1.56 & 1.02 \\
        gpt-4-turbo    & 1.06  & 1.02 & 1.03 & 1.06 &  & 1.02  & 0.93 & 1.13 & 1.08 &  & 0.96 & 1.22 & 1.04 & 1.12 \\
        claude-3-opus  & 1.06  & 1.07 & 1.03 & 1.02 &  & 0.94  & 0.99 & 1.07 & 0.98 &  & 1.05 & 1.12 & 1.10 & 1.02 \\
        gemini-1.5-pro & 0.99  & 1.03 & 0.98 & 0.99 &  & 1.00  & 0.99 & 0.84 & 0.90 &  & 0.88 & 1.11 & 1.23 & 1.02 \\
        mistral-large  & 1.15  & 1.06 & 1.05 & 1.01 &  & 1.38  & 1.03 & 1.09 & 1.09 &  & 1.06 & 1.20 & 1.20 & 1.14 \\ 
        \hline\hline
    \end{tabular}}
\end{table}

{\footnotesize
\linespread{0.5}\selectfont
\begin{table}[htbp]
    \centering
    \caption{Correlations between Inverse Probability, Posterior and Similarity}
    \label{table:corr2}
    \begin{tabular}{l|cc|cc}
        \hline
        \hline
         & \multicolumn{2}{c|}{\textbf{Inverse vs. Posterior}} & \multicolumn{2}{c}{\textbf{Inverse vs. Similarity}} \\ 
        \hline
        Model Name     & Spearman & Pearson & Spearman & Pearson \\ 
        \hline
        gpt-4o         & 0.93***  & 0.90*** & 0.94***  & 0.97*** \\
        gpt-4-turbo    & 0.97***  & 0.98*** & 0.92***  & 0.98*** \\
        claude-3-opus  & 0.96***  & 0.96*** & 0.93***  & 0.98*** \\
        gemini-1.5-pro & 0.95***  & 0.97*** & 0.92***  & 0.92*** \\
        mistral-large  & 0.98***  & 0.98*** & 0.97***  & 0.98*** \\ 
        \hline
        \hline
		\multicolumn{5}{p{0.6\textwidth}}{{\footnotesize Note: two-sided Fisher's p-value, significance levels: ***$p < 0.001$, **$p < 0.01$, *$p < 0.05$.}}
    \end{tabular}
\end{table}

\begin{table}[htbp]
    \centering
    \caption{GPT-4o's Posterior Probabilities in The Semi-structured Test with Prompt Modification}
    \label{tab:semi_modify}
    \begin{tabular}{l|cccc}
        \hline
        \hline
        \textbf{} &
          \textbf{Original} &
          \textbf{\begin{tabular}[c]{@{}c@{}}indicate\\ $\downarrow$\\ compute\end{tabular}} &
          \textbf{\begin{tabular}[c]{@{}c@{}}probability\\ $\downarrow$\\ posterior probability\end{tabular}} &
          \textbf{\begin{tabular}[c]{@{}c@{}}probability\\ $\downarrow$\\ posterior\end{tabular}} \\ 
        \hline
        \multicolumn{5}{c}{\textbf{Uncharacteristic Description}}                    \\ 
        \hline
        High base rate (75\%) posterior    & 0.75    & 0.75    & 0.75     & 0.75     \\
        Low base rate (25\%) posterior     & 0.25    & 0.25    & 0.25     & 0.25     \\
        Difference in posteriors           & 0.5     & 0.49    & 0.5      & 0.5      \\
        Ratio of posterior odds            & 9       & 8.69    & 9        & 9        \\ 
        \hline
        \multicolumn{5}{c}{\textbf{Representative Description --- Computer Science}} \\ 
        \hline
        High base rate (75\%) posterior    & 0.78    & 0.95    & 0.95     & 0.95     \\
        Low base rate (25\%) posterior     & 0.74    & 0.67    & 0.66     & 0.66     \\
        Difference in posteriors           & 0.04    & 0.28    & 0.28     & 0.29     \\
        Ratio of posterior odds            & 1.25    & 9.98    & 8.79     & 9.48     \\ 
        \hline
        \multicolumn{5}{c}{\textbf{Representative Description --- Humanities}}       \\ 
        \hline
        High base rate (75\%) posterior    & 0.24    & 0.41    & 0.39     & 0.39     \\
        Low base rate (25\%) posterior     & 0.24    & 0.08    & 0.05     & 0.06     \\
        Difference in posteriors           & 0.01    & 0.33    & 0.34     & 0.33     \\
        Ratio of posterior odds            & 1.04    & 8.51    & 11.71    & 10.79    \\ 
        \hline
        \hline
    \end{tabular}
\end{table}

\begin{table}[htbp]
\centering
\caption{Correlations between GPT-4o's Prior, Posterior and Similarity with Prompt Modifications}
\label{tab:un_modify}
\begin{tabular}{lccccc}
        \hline
        \hline
                         & \multicolumn{2}{c}{\textbf{Prior vs. Posterior}}     & \multicolumn{2}{c}{\textbf{Posterior vs. Similarity}} \\ 
        \hline
        Prompt Type    & Spearman & Pearson  & Spearman & Pearson \\ 
        \hline
        Original       & -0.88*** & -0.95*** & 0.93***  & 0.90*** \\
        Modification 1 & -0.94*** & -0.99*** & 0.92***  & 0.95*** \\
        Modification 2 & -0.94*** & -0.94*** & 0.93***  & 0.97*** \\
        Modification 3 & -0.87*** & -0.91*** & 0.88***  & 0.96*** \\
        \textit{\textbf{Modification 4}} & \textit{\textbf{-0.75**}} & \textit{\textbf{-0.75**}} & \textit{\textbf{0.87***}}  & \textit{\textbf{0.85**}}  \\ 
        \hline
        \hline
        \multicolumn{5}{p{0.6\textwidth}}{{\footnotesize Note: two-sided Fisher's p-value, significance levels: ***$p < 0.001$, **$p < 0.01$, *$p < 0.05$.}}
\end{tabular}
\end{table}

\newpage

\section*{Supplementary Information}

\section{Details of Empirical Results}

In the structured and unstructured test, results were presented in a somewhat aggregated way. Herein, we provide the details of these findings. The results of the semi-structured test have been documented in the paper, and therefore, are not included here.

\subsection{Structured Test}

In the following tables, a blank accuracy rate indicates that the model either declined to provide a definitive answer or failed to deliver the response in a consistent and parsable format. In such instances, the accuracy can be regarded as zero.

\begin{table}[htbp]
\centering
\caption{Unstructured Tests: Accuracy Rates}
\label{table1}
\resizebox{\columnwidth}{!}{%
\begin{tabular}{lcccccccccccccccc}
\hline
 &
  \textbf{Q1} &
  \textbf{Q2} &
  \textbf{Q3} &
  \textbf{Q4} &
  \textbf{Q5} &
  \textbf{Q6} &
  \textbf{Q7} &
  \textbf{Q8} &
  \textbf{Q9} &
  \textbf{Q10} &
  \textbf{Q11} &
  \textbf{Q12} &
  \textbf{Q13} &
  \textbf{Q14} &
  \textbf{Q15} &
  \textbf{Q16} \\ \hline
gpt-4o                   & 0.97 & 0.99 & 0.92 & 1    & 0.94 & 0.43 & 1    & 0.72 & 0.82 & 0.33 & 1    & 1    & 1    & 1    & 1    & 1    \\
gpt-4-turbo              & 0.86 & 0.99 & 1    & 1    & 0.97 & 0.99 & 1    & 0.67 & 1    & 0.31 & 1    & 1    & 1    & 1    & 1    & 1    \\
gpt-3.5-turbo-0125       & 0.02 & 0.25 & 1    & 0    & 0.47 & 0.37 & 0.02 & 0.06 & 0.22 & 0.01 & 0.13 & 0.06 & 0.99 & 0.87 & 0.96 & 0.99 \\
gemini-1.5-pro-latest    & 0.97 & 0.91 & 0.94 & 1    & 0.96 & 0.59 & 0.86 & 0.91 & 0.89 & 0.82 &      & 0.71 & 1    & 1    & 1    & 1    \\
gemini-1.5-flash-latest  & 0.45 & 0.7  & 0.76 & 0    & 0    & 0.83 & 0.71 & 0.66 & 1    & 0.78 & 0.85 & 0.77 & 1    & 1    & 1    & 1    \\
gemini-1.0-pro-latest    & 0.15 & 0.05 & 0.88 & 0.02 & 0.27 & 0.26 & 0.02 & 0.11 & 0.08 & 0.03 & 0.09 & 0.04 & 0.98 & 0.99 & 0.99 & 0.99 \\
claude-3-opus-20240229   & 0.96 & 0.76 & 0.92 & 1    & 1    & 0.5  & 1    & 0.87 & 1    & 0.2  & 1    & 1    & 1    & 1    & 1    & 1    \\
claude-3-sonnet-20240229 & 0.77 & 0.79 & 0.7  & 0.4  & 0.39 & 0.6  & 0.84 & 0.85 & 0.93 & 0.05 & 0.81 & 0.91 & 1    & 1    & 1    & 1    \\
claude-3-haiku-20240307  & 0.66 & 0.26 & 0.52 & 0.31 & 0.3  & 0.04 & 0.24 & 0.36 & 0.9  & 0.1  & 0.25 & 0.39 & 0.91 & 0.99 & 0.99 & 1    \\
open-mixtral-8x22b       &      & 0.61 & 0.87 & 0.02 & 0.97 & 0.91 & 0.63 & 0.58 & 0.96 & 0.3  & 0.65 & 0.69 & 1    & 1    & 1    & 1    \\
open-mixtral-8x7b        & 0.02 & 0.2  & 0.68 & 0    & 1    & 0    & 0.27 & 0.15 & 0.7  & 0.1  & 0.22 & 0.37 & 1    & 1    & 1    & 0.02 \\
open-mistral-7b          & 0.42 & 0.03 & 0.42 &      &      &      &      &      &      &      & 0.02 & 0.01 &      &      &      &      \\
mistral-large-latest     & 1    & 1    & 0.93 & 1    & 1    & 0.99 & 1    & 0.68 & 1    & 0.26 & 1    & 1    & 1    & 1    & 1    & 1    \\
mistral-medium-latest    & 0.66 & 0.51 & 0.33 & 0    & 1    & 1    & 0.48 & 0.2  & 1    & 0.22 & 0.55 & 0.41 &      &      &      &      \\
mistral-small-latest     &      & 0.3  & 0.12 & 0.48 & 0.71 & 0.78 & 0.03 & 0.1  & 1    & 0.03 & 0.03 & 0.03 & 0.36 & 1    & 0.99 & 1    \\ \hline
\end{tabular}%
}
\end{table}
\FloatBarrier

\subsection{Unstructured Test}

$S(A)$, $S(B)$ and $S(C)$ denote the similarity scores assigned to fields $A,B$ and $C$, respectively 

%
\begin{longtable}[c]{lllllllllllllll}
\caption{Unstructured Test: Sample Mean of Model Responses}
\label{table2}\\
\hline
 &
  \textbf{A} &
  \textbf{AB} &
  \textbf{AC} &
  \textbf{ABC} &
   &
  \textbf{B} &
  \textbf{AB} &
  \textbf{BC} &
  \textbf{ABC} &
   &
  \textbf{C} &
  \textbf{AC} &
  \textbf{BC} &
  \textbf{ABC} \\ \hline
\endfirsthead
\multicolumn{15}{c}%
{{\bfseries Table \thetable\ continued from previous page}} \\
\endhead
\hline
\endfoot
\endlastfoot
\multicolumn{15}{c}{\textbf{gpt-4o}} \\ \hline
$P(A)$ &
  0.02 &
  0.03 &
  0.03 &
  \multicolumn{1}{l|}{0.04} &
  $P(B)$ &
  0.17 &
  0.19 &
  0.19 &
  \multicolumn{1}{l|}{0.2} &
  $P(C)$ &
  0.13 &
  0.14 &
  0.13 &
  0.13 \\
$P(E|A)$ &
  0.81 &
  0.84 &
  0.76 &
  \multicolumn{1}{l|}{0.75} &
  $P(E|B)$ &
  0.2 &
  0.16 &
  0.23 &
  \multicolumn{1}{l|}{0.14} &
  $P(E|C)$ &
  0.44 &
  0.36 &
  0.72 &
  0.39 \\
$P(E|\neg A)$ &
  0.75 &
  0.8 &
  0.75 &
  \multicolumn{1}{l|}{0.74} &
  $P(E|\neg B)$ &
  0.15 &
  0.14 &
  0.15 &
  \multicolumn{1}{l|}{0.13} &
  $P(E|\neg C)$ &
  0.3 &
  0.29 &
  0.46 &
  0.38 \\
$P(A|E)$ &
  0.44 &
  0.7 &
  0.66 &
  \multicolumn{1}{l|}{0.62} &
  $P(B|E)$ &
  0.1 &
  0.11 &
  0.1 &
  \multicolumn{1}{l|}{0.09} &
  $P(C|E)$ &
  0.19 &
  0.27 &
  0.29 &
  0.24 \\
$S(A)$ &
  0.87 &
  0.9 &
  0.89 &
  \multicolumn{1}{l|}{0.9} &
  $S(B)$ &
  0.29 &
  0.24 &
  0.32 &
  \multicolumn{1}{l|}{0.26} &
  $S(C)$ &
  0.68 &
  0.53 &
  0.71 &
  0.58 \\ \hline
\multicolumn{15}{c}{\textbf{gpt-4-turbo}} \\ \hline
$P(A)$ &
  0.02 &
  0.03 &
  0.04 &
  \multicolumn{1}{l|}{0.04} &
  $P(B)$ &
  0.2 &
  0.23 &
  0.24 &
  \multicolumn{1}{l|}{0.23} &
  $P(C)$ &
  0.09 &
  0.14 &
  0.17 &
  0.12 \\
$P(E|A)$ &
  0.77 &
  0.8 &
  0.76 &
  \multicolumn{1}{l|}{0.77} &
  $P(E|B)$ &
  0.27 &
  0.22 &
  0.3 &
  \multicolumn{1}{l|}{0.21} &
  $P(E|C)$ &
  0.52 &
  0.46 &
  0.68 &
  0.46 \\
$P(E|\neg A)$ &
  0.73 &
  0.78 &
  0.74 &
  \multicolumn{1}{l|}{0.73} &
  $P(E|\neg B)$ &
  0.26 &
  0.24 &
  0.27 &
  \multicolumn{1}{l|}{0.2} &
  $P(E|\neg C)$ &
  0.54 &
  0.37 &
  0.65 &
  0.41 \\
$P(A|E)$ &
  0.68 &
  0.64 &
  0.72 &
  \multicolumn{1}{l|}{0.64} &
  $P(B|E)$ &
  0.14 &
  0.17 &
  0.2 &
  \multicolumn{1}{l|}{0.14} &
  $P(C|E)$ &
  0.3 &
  0.35 &
  0.58 &
  0.37 \\
$S(A)$ &
  0.8 &
  0.82 &
  0.84 &
  \multicolumn{1}{l|}{0.87} &
  $S(B)$ &
  0.29 &
  0.32 &
  0.37 &
  \multicolumn{1}{l|}{0.34} &
  $S(C)$ &
  0.64 &
  0.61 &
  0.73 &
  0.63 \\ \hline
\multicolumn{15}{c}{\textbf{gemini-1.5-pro-latest}} \\ \hline
$P(A)$ &
  0.05 &
  0.05 &
  0.04 &
  \multicolumn{1}{l|}{0.05} &
  $P(B)$ &
  0.2 &
  0.3 &
  0.32 &
  \multicolumn{1}{l|}{0.3} &
  $P(C)$ &
  0.2 &
  0.17 &
  0.18 &
  0.17 \\
$P(E|A)$ &
  0.73 &
  0.73 &
  0.67 &
  \multicolumn{1}{l|}{0.71} &
  $P(E|B)$ &
  0.27 &
  0.26 &
  0.31 &
  \multicolumn{1}{l|}{0.14} &
  $P(E|C)$ &
  0.35 &
  0.37 &
  0.65 &
  0.37 \\
$P(E|\neg A)$ &
  0.74 &
  0.71 &
  0.69 &
  \multicolumn{1}{l|}{0.71} &
  $P(E|\neg B)$ &
  0.27 &
  0.26 &
  0.37 &
  \multicolumn{1}{l|}{0.16} &
  $P(E|\neg C)$ &
  0.39 &
  0.34 &
  0.53 &
  0.37 \\
$P(A|E)$ &
  0.71 &
  0.68 &
  0.69 &
  \multicolumn{1}{l|}{0.65} &
  $P(B|E)$ &
  0.14 &
  0.11 &
  0.16 &
  \multicolumn{1}{l|}{0.11} &
  $P(C|E)$ &
  0.33 &
  0.32 &
  0.48 &
  0.31 \\
$S(A)$ &
  0.8 &
  0.77 &
  0.78 &
  \multicolumn{1}{l|}{0.8} &
  $S(B)$ &
  0.4 &
  0.32 &
  0.35 &
  \multicolumn{1}{l|}{0.35} &
  $S(C)$ &
  0.6 &
  0.58 &
  0.7 &
  0.63 \\ \hline
\multicolumn{15}{c}{\textbf{claude-3-opus-20240229}} \\ \hline
$P(A)$ &
  0.02 &
  0.02 &
  0.02 &
  \multicolumn{1}{l|}{0.02} &
  $P(B)$ &
  0.22 &
  0.19 &
  0.19 &
  \multicolumn{1}{l|}{0.2} &
  $P(C)$ &
  0.1 &
  0.09 &
  0.09 &
  0.09 \\
$P(E|A)$ &
  0.75 &
  0.77 &
  0.74 &
  \multicolumn{1}{l|}{0.75} &
  $P(E|B)$ &
  0.28 &
  0.26 &
  0.32 &
  \multicolumn{1}{l|}{0.21} &
  $P(E|C)$ &
  0.54 &
  0.42 &
  0.69 &
  0.43 \\
$P(E|\neg A)$ &
  0.71 &
  0.72 &
  0.72 &
  \multicolumn{1}{l|}{0.74} &
  $P(E|\neg B)$ &
  0.29 &
  0.26 &
  0.3 &
  \multicolumn{1}{l|}{0.22} &
  $P(E|\neg C)$ &
  0.51 &
  0.38 &
  0.63 &
  0.42 \\
$P(A|E)$ &
  0.7 &
  0.63 &
  0.63 &
  \multicolumn{1}{l|}{0.53} &
  $P(B|E)$ &
  0.18 &
  0.16 &
  0.16 &
  \multicolumn{1}{l|}{0.1} &
  $P(C|E)$ &
  0.32 &
  0.27 &
  0.43 &
  0.26 \\
$S(A)$ &
  0.8 &
  0.8 &
  0.8 &
  \multicolumn{1}{l|}{0.83} &
  $S(B)$ &
  0.39 &
  0.3 &
  0.4 &
  \multicolumn{1}{l|}{0.31} &
  $S(C)$ &
  0.6 &
  0.6 &
  0.77 &
  0.6 \\ \hline
\multicolumn{15}{c}{\textbf{mistral-large-latest}} \\ \hline
$P(A)$ &
  0.02 &
  0.03 &
  0.03 &
  \multicolumn{1}{l|}{0.03} &
  $P(B)$ &
  0.25 &
  0.25 &
  0.25 &
  \multicolumn{1}{l|}{0.25} &
  $P(C)$ &
  0.14 &
  0.15 &
  0.15 &
  0.15 \\
$P(E|A)$ &
  0.83 &
  0.75 &
  0.74 &
  \multicolumn{1}{l|}{0.73} &
  $P(E|B)$ &
  0.3 &
  0.26 &
  0.3 &
  \multicolumn{1}{l|}{0.2} &
  $P(E|C)$ &
  0.63 &
  0.49 &
  0.7 &
  0.46 \\
$P(E|\neg A)$ &
  0.72 &
  0.71 &
  0.71 &
  \multicolumn{1}{l|}{0.72} &
  $P(E|\neg B)$ &
  0.22 &
  0.25 &
  0.28 &
  \multicolumn{1}{l|}{0.19} &
  $P(E|\neg C)$ &
  0.6 &
  0.41 &
  0.58 &
  0.41 \\
$P(A|E)$ &
  0.76 &
  0.7 &
  0.7 &
  \multicolumn{1}{l|}{0.61} &
  $P(B|E)$ &
  0.12 &
  0.16 &
  0.2 &
  \multicolumn{1}{l|}{0.12} &
  $P(C|E)$ &
  0.42 &
  0.52 &
  0.6 &
  0.36 \\
$S(A)$ &
  0.85 &
  0.81 &
  0.8 &
  \multicolumn{1}{l|}{0.8} &
  $S(B)$ &
  0.4 &
  0.35 &
  0.35 &
  \multicolumn{1}{l|}{0.33} &
  $S(C)$ &
  0.6 &
  0.6 &
  0.7 &
  0.6 \\ \hline
\end{longtable}
\FloatBarrier

\begin{table}[htbp]
\centering
\caption{Unstructured Test Prompt Engineering: Sample Mean of GPT-4o Responses}
\label{table3}
\resizebox{\columnwidth}{!}{%
\begin{tabular}{lcccc|lcccc|lcccc}
\hline
               & \multicolumn{4}{c|}{P(A|E)} &                & \multicolumn{4}{c|}{P(B|E)} &                & \multicolumn{4}{c}{P(C|E)} \\ \hline
               & A     & AB    & AC   & ABC  &                & B     & AB    & BC   & ABC  &                & C     & AC   & BC   & ABC  \\ \hline
Modification 1 & 0.75  & 0.72  & 0.69 & 0.65 & Modification 1 & 0.15  & 0.1   & 0.11 & 0.08 & Modification 1 & 0.31  & 0.27 & 0.37 & 0.25 \\
Modification 2 & 0.74  & 0.79  & 0.7  & 0.68 & Modification 2 & 0.12  & 0.13  & 0.21 & 0.11 & Modification 2 & 0.44  & 0.29 & 0.62 & 0.31 \\
Modification 3 & 0.61  & 0.71  & 0.61 & 0.59 & Modification 3 & 0.1   & 0.15  & 0.21 & 0.11 & Modification 3 & 0.36  & 0.31 & 0.61 & 0.32 \\
Modification 4 & 0.26  & 0.42  & 0.31 & 0.39 & Modification 4 & 0.04  & 0.09  & 0.11 & 0.09 & Modification 4 & 0.16  & 0.25 & 0.41 & 0.3  \\ \hline
\end{tabular}%
}
\end{table}

\section{Prompt Details}

We now present the detailed experimental prompts for each question in all three main tests. Except for the random shuffle of the order, the prompts here are identical in format and content to those used in the experiment.

\subsection{Structured Test}

For the structured test prompts, the correct answer is specified in the title.

\begin{promptbox}[Q1 --- Answer: 45]
\linespread{0.5}\selectfont
All families of six children in a city were surveyed. In 45 families, the exact order of births of boys (B) and girls (G) was G B G B B G.
\vspace{5pt}

What is your estimate of the number of families surveyed in which the exact order of births was B G B B B B?
\vspace{5pt}

Please answer the question in the following format: \{answer: your estimate\}
\end{promptbox}
\FloatBarrier

\begin{promptbox}[Q2 --- Answer: 4]
\linespread{0.5}\selectfont
Consider two very large decks of cards, denoted A and B. In deck A, 2/3 of the cards are marked X, and 1/3 are marked O. In deck B, 1/3 of the cards are marked X, and 2/3 are marked O.
\vspace{5pt}

One of the decks has been selected by chance, and 4 cards have been drawn at random from it, of which 3 are marked X and 1 are marked O.
\vspace{5pt}

What are the posterior odds that the 4 cards were drawn from Deck A, the deck where most cards are marked X, compared to Deck B, the deck where most cards are marked O?
\vspace{5pt}

Please answer the question in the following format: \{odds: your answer as a single number\}
\end{promptbox}
\FloatBarrier

\begin{promptbox}[Q3 --- Answer: 1]
\linespread{0.5}\selectfont
One each round of a game, 6 distinguishable balls are distributed at random among three children: A, B, C.
Consider the following two cases of how the balls are distributed:
\vspace{5pt}

case 1 --- A:2, B:2, C:2

case 2 --- A:2, B:3, C:1
\vspace{5pt}

In many rounds of the game, will there be more results of case 1 or case 2? Please feel free to compute and reason, and answer the question in the following format: \{case: your answer, 1 or 2\}
\end{promptbox}
\FloatBarrier

\begin{promptbox}[Q4 --- Answer: 2]
\linespread{0.5}\selectfont
A certain town is served by two hospitals. In hospital 1 about 45 babies are born each day, and in hospital 2 about 15
babies are born each day.
\vspace{5pt}

About 50\% of all babies are boys, but the exact percentage of baby boys varies from day to day. Sometimes it may be
higher than 50\%, sometimes lower.
\vspace{5pt}

For a period of 1 year, each hospital recorded the days on which more than 60\% of the babies born were boys.
\vspace{5pt}

Which hospital do you think recorded more such days? Or do you think they are about the same (within 5\% of each other)?
\vspace{5pt}

Please answer 1 if you think hospital 1 recorded more, 2 if you think hospital 2 recorded more, and 3 if you think they are about the same (within 5\% of each other).
\vspace{5pt}

Please answer the question in the following format: \{hospital: your answer\}
\end{promptbox}
\FloatBarrier

\begin{promptbox}[Q5 --- Answer: 2]
\linespread{0.5}\selectfont
There are two investigators selected a paperback to study some properties of language.
\vspace{5pt}

Investigator 1 computed the average word-length in every page of the book. Investigator 2 took the first line in each
page and computed the line's average word-length.
\vspace{5pt}

The average word-length in the entire book is 4. However, not every line or page has exactly that average. Some may
have a higher average word-length, some lower.
\vspace{5pt}

Investigator 1 counted the number of pages that had an average word-length of 6 or more and Investigator 2 counted
the number of lines that had an average word-length of 6 or more.
\vspace{5pt}

Which investigator do you think recorded a larger number of such units? Or do you think they are about the same
(within 5\% of each other)?
\vspace{5pt}

Please answer 1 if you think investigator 1 recorded more, 2 if you think investigator 2 recorded more, and 3 if you think they are about the same (within 5\% of each other).
\vspace{5pt}

Please answer the question in the following format: \{investigator: your answer\}
\end{promptbox}
\FloatBarrier

\begin{promptbox}[Q6 --- Answer: 2]
\linespread{0.5}\selectfont
A medical survey is being held to study some factors pertaining to a certain disease. Two teams are collecting data
from a population. The average height of men in the population is around 178 cm, and there are as many men whose height
is above average as there are men whose height is below average.
\vspace{5pt}

In a period of 365 days, each day Team 1 randomly surveys three men from the population. During the survey, the team
ranks the three men with respect to their height. At the end of the period, the team counts the total number of days
on which the height of the middle man is more than 183 cm.

\vspace{5pt}

In the same period, each day Team 2 randomly surveys only one man from the population. At the end of the period, the
team simply counts the total number of days on which the height of the man they surveyed was taller than 183 cm.
\vspace{5pt}

Which team do you think counted more such days? Or do you think they are about the same (within 5\% of each other)?
\vspace{5pt}

Please answer 1 if you think team 1 recorded more, 2 if you think team 2 recorded more, and 3 if you think they are
about the same (within 5\% of each other).
\vspace{5pt}

Please answer the question in the follow format:
\{team: your answer\}
\end{promptbox}
\FloatBarrier

The responses to Q7, Q8, and Q10-Q12 are approximately $0.4138$. Taking into account potential computational challenges and the varying decimal precision that LLMs might produce, we consider any answer within the interval $(0.40, 0.43)$ as correct.

\begin{promptbox}[Q7]
\linespread{0.5}\selectfont
A person was attacked from behind at night and the victim could not see the attacker's height.
Despite this, we have the following two available information:
\vspace{5pt}

a. 85\% of the population in the city is shorter than 185 cm, and 15\% are 185 cm or taller.

b. A witness identified the attacker as being 185 cm or taller. The court assessed the witness's ability to judge a
person's height under conditions similar to that night. Given a sample of people, half shorter than 185 cm and
half 185 cm or taller, the witness correctly identified height 80\% of the time and erred in 20\% of the cases.
\vspace{5pt}

What is your estimate of the probability that the attacker's height was 185 cm or taller?
\vspace{5pt}

Please answer the question in the following format:
\{probability: your estimate, please provide a number between 0 and 1\}
\end{promptbox}
\FloatBarrier

\begin{promptbox}[Q8]
\linespread{0.5}\selectfont
A person was attacked from behind at night and the victim could not see the attacker's height.
Despite this, we have the following two available information:
\vspace{5pt}

a. The population of people shorter than 185 cm and those 185 cm and taller is roughly equal in size.
However, 85\% of such attackers are under 185 cm, while only 15\% attackers are 185 cm or taller.

b. A witness identified the attacker as being 185 cm or taller. The court assessed the witness's ability to judge a
person's height under conditions similar to that night. Given a sample of people, half shorter than 185 cm and
half 185 cm or taller, the witness correctly identified height 80\% of the time and erred in 20\% of the cases.
\vspace{5pt}

What is your estimate of the probability that the attacker's height was 185 cm or taller?
\vspace{5pt}

Please answer the question in the following format:
\{probability: your estimate, please provide a number between 0 and 1\}
\end{promptbox}
\FloatBarrier

\begin{promptbox}[Q9 --- Answer: 0.2]
\linespread{0.5}\selectfont
A person was attacked from behind at night and the victim could not see the attacker's height.
In the city, 85\% of the population is shorter than 185 cm, and 15\% are 185 cm or taller.
\vspace{5pt}

The police investigation discovered that in the neighborhood in which the attack occurred, 80\% of the people are
shorter than 185 cm and 20\% are 185 cm or taller.
\vspace{5pt}

What is your estimate of the probability that the attacker's height was 185 cm or taller?
\vspace{5pt}

Please answer the question in the following format:
\{probability: your estimate, please provide a number between 0 and 1\}
\end{promptbox}
\FloatBarrier

\begin{promptbox}[Q10]
\linespread{0.5}\selectfont
A person was attacked from behind at night and the victim could not see the attacker's height.
Despite this, we have the following two available information:
\vspace{5pt}

a. 85\% of the population in the city is shorter than 185 cm, and 15\% are 185 cm or taller.

b. There were two witnesses to the accident. One claimed that the attacker had been 185 cm or taller, and the other
claimed that the attacker had been shorter than 185 cm. The court assessed the witness's ability to judge a
person's height under conditions similar to that night. Given a sample of people, half shorter than 185 cm and
half 185 cm or taller, the first witness was able to identify the correct height about 80\% of the time and erred in
20\% of the cases. The second witness was able to identify the correct height about 70\% of the time and erred in 30\% of the cases.
\vspace{5pt}

What is your estimate of the probability that the attacker's height was 185 cm or taller, as the first witness claimed?
\vspace{5pt}

Please answer the question in the following format:
\{probability: your estimate, please provide a number between 0 and 1\}
\end{promptbox}
\FloatBarrier

\begin{promptbox}[Q11]
\linespread{0.5}\selectfont
A person was attacked from behind at night, and the victim could not see the attacker's race. Despite this, we have
the following two pieces of information:
\vspace{5pt}

a. 85\% of the population in the city is white, and 15\% is black.

b. A witness identified the attacker as black. The court assessed the witness's ability to judge a person's race under
conditions similar to that night. Given a sample of people, half white and half black, the witness correctly identified
race 80\% of the time and erred in 20\% of the cases.
\vspace{5pt}

What is your estimate of the probability that the attacker is black?
\vspace{5pt}

Please answer the question in the following format: {probability: your estimate, please provide a
number between 0 and 1}
\end{promptbox}
\FloatBarrier

\begin{promptbox}[Q12]
\linespread{0.5}\selectfont
A person was attacked from behind at night, and the victim could not see the attacker's gender. Despite this, we have
the following two pieces of information:
\vspace{5pt}

a. 85\% of the population in the city is female, and 15\% is male.

b. A witness identified the attacker as male. The court assessed the witness's ability to judge a person's gender under
conditions similar to that night. Given a sample of people, half female and half male, the witness correctly identified
gender 80\% of the time and erred in 20\% of the cases.
\vspace{5pt}

What is your estimate of the probability that the attacker is male?
\vspace{5pt}

Please answer the question in the following format: \{probability: your estimate, please provide a
number between 0 and 1\}
\end{promptbox}
\FloatBarrier

\begin{promptbox}[Q13 --- Answer: 0.2]
\linespread{0.5}\selectfont
Studies of dreaming have shown that 80\% of people of both sexes report that they dream, if only occasionally,
whereas 20\% claim they do not remember ever dreaming. Accordingly, people are classified by dream investigators
as "Dreamers" or "Nondreamers". With respect to dreaming, mating is completely random.
\vspace{5pt}

Brett is a Nondreamer.
\vspace{5pt}

What do you think are the probabilities that his wife is also a Nondreamer?
\vspace{5pt}

Please answer the question in the following format:
\{probability: your answer, please provide a number between 0 and 1\}
\end{promptbox}
\FloatBarrier

\begin{promptbox}[Q14 --- Answer: 0.2]
\linespread{0.5}\selectfont
Studies of dreaming have shown that 80\% of people of both sexes report that they dream, if only occasionally,
whereas 20\% claim they do not remember ever dreaming. Accordingly, people are classified by dream investigators
as "Dreamers" or "Nondreamers". With respect to dreaming, the classification of husband and wife was found to be
independent.
\vspace{5pt}

Brett is a Nondreamer.
\vspace{5pt}

What do you think are the probabilities that his wife is also a Nondreamer?
\vspace{5pt}

Please answer the question in the following format:
\{probability: your answer, please provide a number between 0 and 1\}
\end{promptbox}
\FloatBarrier

\begin{promptbox}[Q15 --- Answer: 0.2]
\linespread{0.5}\selectfont
Studies of dreaming have shown that 80\% of people of both sexes report that they dream, if only occasionally,
whereas 20\% claim they do not remember ever dreaming. Accordingly, people are classified by dream investigators
as "Dreamers" or "Nondreamers". The study has also found that the spouse's classification has no predictive validity
regarding one's classification about dreaming.
\vspace{5pt}

Brett is a Nondreamer.
\vspace{5pt}

What do you think are the probabilities that his wife is also a Nondreamer?
\vspace{5pt}

Please answer the question in the following format:
\{probability: your answer, please provide a number between 0 and 1\}
\end{promptbox}
\FloatBarrier

\begin{promptbox}[Q16 --- Answer: 0.2]
\linespread{0.5}\selectfont
Studies of dreaming have shown that 80\% of people of both sexes report that they dream, if only occasionally,
whereas 20\% claim they do not remember ever dreaming. Accordingly, people are classified by dream investigators
as "Dreamers" or "Nondreamers". With respect to dreaming, the classification is found to be independent in successive
generations.
\vspace{5pt}

Brett is a Nondreamer.
\vspace{5pt}

What do you think are the probabilities that his father is also a Nondreamer?
\vspace{5pt}

Please answer the question in the following format:
\{probability: your answer, please provide a number between 0 and 1\}
\end{promptbox}
\FloatBarrier

\subsection{Semi-Structured Test}

\

\begin{promptbox}[Uncharacteristic Description --- High Base Rate]
\linespread{0.5}\selectfont
A study aimed to identify common personality traits among successful academics was conducted. In this study, a panel of professional psychologists interviewed and administered personality tests to 75 tenured and successful professors in computer science and 25 tenured and successful professors in the humanities. Based on these personality tests, descriptions were written for each of the 100 professors.
\vspace{5pt}

You will be presented with one such description, chosen at random from the 100 available descriptions.
\vspace{5pt}

Here is the description:
Jason is a 30-year-old man, married and without children. He is known for his high ability and strong motivation. Well-liked by his colleagues, Jason enjoys a positive reputation in his professional environment.
\vspace{5pt}

Now, please indicate your probability that Jason is one of the computer science professors in the sample.
\vspace{5pt}

Please give a number between 0 and 1, and answer the question in the following format:
\vspace{5pt}

\{"probability": "your answer"\}
\end{promptbox}
\FloatBarrier

\begin{promptbox}[Uncharacteristic Description --- Low Base Rate]
\linespread{0.5}\selectfont
A study aimed to identify common personality traits among successful academics was conducted. In this study, a panel of professional psychologists interviewed and administered personality tests to 25 tenured and successful professors in computer science and 75 tenured and successful professors in the humanities. Based on these personality tests, descriptions were written for each of the 100 professors.
\vspace{5pt}

You will be presented with one such description, chosen at random from the 100 available descriptions.
\vspace{5pt}

Here is the description:
Jason is a 30-year-old man, married and without children. He is known for his high ability and strong motivation. Well-liked by his colleagues, Jason enjoys a positive reputation in his professional environment.
\vspace{5pt}

Now, please indicate your probability that Jason is one of the computer science professors in the sample.
\vspace{5pt}

Please give a number between 0 and 1, and answer the question in the following format:
\vspace{5pt}

\{"probability": "your answer"\}
\end{promptbox}
\FloatBarrier

\begin{promptbox}[Representative Description of Computer Science --- High Base Rate]
\linespread{0.5}\selectfont
A study aimed to identify common personality traits among successful academics was conducted. In this study, a panel of professional psychologists interviewed and administered personality tests to 75 tenured and successful professors in computer science and 25 tenured and successful professors in the humanities. Based on these personality tests, descriptions were written for each of the 100 professors.
\vspace{5pt}

You will be presented with one such description, chosen at random from the 100 available descriptions.
\vspace{5pt}

Here is the description:
Jason is a 45-year-old married man with five children. He is conservative, careful, talented, and ambitious. Jason enjoys coding, solving mathematical puzzles, and jogging. Despite being highly creative, he does not enjoy drawing or writing.
\vspace{5pt}

Now, please indicate your probability that Jason is one of the computer science professors in the sample.
\vspace{5pt}

Please give a number between 0 and 1, and answer the question in the following format:
\vspace{5pt}

\{"probability": "your answer"\}
\end{promptbox}
\FloatBarrier

\begin{promptbox}[Representative Description of Computer Science --- Low Base Rate]
\linespread{0.5}\selectfont
A study aimed to identify common personality traits among successful academics was conducted. In this study, a panel of professional psychologists interviewed and administered personality tests to 25 tenured and successful professors in computer science and 75 tenured and successful professors in the humanities. Based on these personality tests, descriptions were written for each of the 100 professors.
\vspace{5pt}

You will be presented with one such description, chosen at random from the 100 available descriptions.
\vspace{5pt}

Here is the description:
Jason is a 45-year-old married man with five children. He is conservative, careful, talented, and ambitious. Jason enjoys coding, solving mathematical puzzles, and jogging. Despite being highly creative, he does not enjoy drawing or writing.
\vspace{5pt}

Now, please indicate your probability that Jason is one of the computer science professors in the sample.
\vspace{5pt}

Please give a number between 0 and 1, and answer the question in the following format:
\vspace{5pt}

\{"probability": "your answer"\}
\end{promptbox}
\FloatBarrier

\begin{promptbox}[Representative Description of Humanities --- High Base Rate]
\linespread{0.5}\selectfont
A study aimed to identify common personality traits among successful academics was conducted. In this study, a panel of professional psychologists interviewed and administered personality tests to 75 tenured and successful professors in computer science and 25 tenured and successful professors in the humanities. Based on these personality tests, descriptions were written for each of the 100 professors.
\vspace{5pt}

You will be presented with one such description, chosen at random from the 100 available descriptions.
\vspace{5pt}

Here is the description:
Jason is a 36-year-old, divorced man with no children. He is creative and imaginative, often emotionally driven. Jason enjoys traveling, reading, and exploring. He is gifted and passionate, though not particularly technical.
\vspace{5pt}

Now, please indicate your probability that Jason is one of the computer science professors in the sample.
\vspace{5pt}

Please give a number between 0 and 1, and answer the question in the following format:
\vspace{5pt}

\{"probability": "your answer"\}
\end{promptbox}
\FloatBarrier

\begin{promptbox}[Representative Description of Humanities --- Low Base Rate]
\linespread{0.5}\selectfont
A study aimed to identify common personality traits among successful academics was conducted. In this study, a panel of professional psychologists interviewed and administered personality tests to 25 tenured and successful professors in computer science and 75 tenured and successful professors in the humanities. Based on these personality tests, descriptions were written for each of the 100 professors.
\vspace{5pt}

You will be presented with one such description, chosen at random from the 100 available descriptions.
\vspace{5pt}

Here is the description:
Jason is a 36-year-old, divorced man with no children. He is creative and imaginative, often emotionally driven. Jason enjoys traveling, reading, and exploring. He is gifted and passionate, though not particularly technical.
\vspace{5pt}

Now, please indicate your probability that Jason is one of the computer science professors in the sample.
\vspace{5pt}

Please give a number between 0 and 1, and answer the question in the following format:
\vspace{5pt}

\{"probability": "your answer"\}
\end{promptbox}
\FloatBarrier

\subsection{Unstructured Test}

Since the unstructured test involves $5$ judgments and each judgment entails $7$ rotations, we believe including all such prompts --- even in the supporting information --- would be excessive. Therefore, for each query, we will only present the prompts for rotation $A$ and rotation $ABC$.

\begin{promptbox}[Prior $P(H)$ --- Rotation A]
\linespread{0.5}\selectfont
Consider students who were enrolled in a graduate program in the U.S. in 2021.
\vspace{5pt}

Please provide your best guesses about the percentage of these students who were enrolled in the following field:
\vspace{5pt}

agricultural and veterinary science
\vspace{5pt}

Please answer the question in the following format:
\vspace{5pt}

\{
  "agricultural and veterinary science": "your answer, please provide a value between 0 and 100"
\}
\end{promptbox}
\FloatBarrier

\begin{promptbox}[Prior $P(H)$ --- Rotation ABC]
\linespread{0.5}\selectfont
Consider students who were enrolled in a graduate program in the U.S. in 2021.
\vspace{5pt}

Please provide your best guesses about the percentage of these students who were enrolled in the following fields:
\vspace{5pt}

computer science

agricultural and veterinary science

business administration
\vspace{5pt}

Your total does not need to add up to 100, as there are other graduate fields of specialization. Please answer the question in the following format:
\vspace{5pt}

\{
  "computer science": "your answer, please provide a value between 0 and 100",

  "agricultural and veterinary science": "your answer, please provide a value between 0 and 100",

  "business administration": "your answer, please provide a value between 0 and 100"
\}
\end{promptbox}
\FloatBarrier

\begin{promptbox}[Posterior $P(H|E)$ --- Rotation A]
\linespread{0.5}\selectfont
The following is a personality sketch of Adam.
\vspace{5pt}

The subject has a genuine curiosity about nature and its various plants. A quiet boy who enjoys solving puzzles, he has a warm heart, strong empathy, and a deep love for animals. He is of high intelligence and is good at understanding abstract concepts. The subject has a strong understanding and appreciation for the hard work and dedication.
\vspace{5pt}

In 2021, Adam was a graduate student in the U.S.
\vspace{5pt}

Please provide your best estimate of the probability that Adam was enrolled in the following field:
\vspace{5pt}

agricultural and veterinary science
\vspace{5pt}

Please answer the question in the following format:
\vspace{5pt}

\{
  "agricultural and veterinary science": "your answer, please provide a value between 0 and 1"
\}
\end{promptbox}
\FloatBarrier

\begin{promptbox}[Posterior $P(H|E)$ --- Rotation ABC]
\linespread{0.5}\selectfont
The following is a personality sketch of Adam.
\vspace{5pt}

The subject has a genuine curiosity about nature and its various plants. A quiet boy who enjoys solving puzzles, he has a warm heart, strong empathy, and a deep love for animals. He is of high intelligence and is good at understanding abstract concepts. The subject has a strong understanding and appreciation for the hard work and dedication.
\vspace{5pt}

In 2021, Adam was a graduate student in the U.S.
\vspace{5pt}

Please provide your best estimate of the probability that Adam was enrolled in the following fields:
\vspace{5pt}

business administration

computer science

agricultural and veterinary science
\vspace{5pt}

Your total does not need to add up to 1, as there are other graduate fields of specialization. Please answer the question in the following format:
\vspace{5pt}

\{
  "business administration": "your answer, please provide a value between 0 and 1",

  "computer science": "your answer, please provide a value between 0 and 1",

  "agricultural and veterinary science": "your answer, please provide a value between 0 and 1"
\}
\end{promptbox}
\FloatBarrier

\begin{promptbox}[Inverse Probability $P(E|H)$ --- Rotation A]
\linespread{0.5}\selectfont
The following is a personality sketch.
\vspace{5pt}

The subject has a genuine curiosity about nature and its various plants. A quiet boy who enjoys solving puzzles, he has a warm heart, strong empathy, and a deep love for animals. He is of high intelligence and is good at understanding abstract concepts. The subject has a strong understanding and appreciation for the hard work and dedication.
\vspace{5pt}

In 2021, Adam was a graduate student in the U.S.
\vspace{5pt}

For the following field, please provide your best estimate of the probability that the personality sketch was about Adam, if Adam was enrolled in that field:
\vspace{5pt}

agricultural and veterinary science
\vspace{5pt}

Please answer the question in the following format:
\vspace{5pt}

\{
  "agricultural and veterinary science": "your answer, please provide a value between 0 and 1"
\}
\end{promptbox}
\FloatBarrier

\begin{promptbox}[Inverse Probability $P(E|H)$  --- Rotation ABC]
\linespread{0.5}\selectfont
The following is a personality sketch.
\vspace{5pt}

The subject has a genuine curiosity about nature and its various plants. A quiet boy who enjoys solving puzzles, he has a warm heart, strong empathy, and a deep love for animals. He is of high intelligence and is good at understanding abstract concepts. The subject has a strong understanding and appreciation for the hard work and dedication.
\vspace{5pt}

In 2021, Adam was a graduate student in the U.S.
\vspace{5pt}

For each of the following fields, please provide your best estimate of the probability that the personality sketch was about Adam, if Adam was enrolled in that field:
\vspace{5pt}

agricultural and veterinary science

business administration

computer science
\vspace{5pt}

Your total does not need to add up to 1. Please answer the question in the following format:
\vspace{5pt}

\{
  "agricultural and veterinary science": "your answer, please provide a value between 0 and 1",

  "business administration": "your answer, please provide a value between 0 and 1",

  "computer science": "your answer, please provide a value between 0 and 1"
\}
\end{promptbox}
\FloatBarrier

\begin{promptbox}[Inverse Probability for Alternative $P(E|\neg H)$ --- Rotation A]
\linespread{0.5}\selectfont
The following is a personality sketch.
\vspace{5pt}

The subject has a genuine curiosity about nature and its various plants. A quiet boy who enjoys solving puzzles, he has a warm heart, strong empathy, and a deep love for animals. He is of high intelligence and is good at understanding abstract concepts. The subject has a strong understanding and appreciation for the hard work and dedication.
\vspace{5pt}

In 2021, Adam was a graduate student in the U.S.
\vspace{5pt}

For the following field, please provide your best estimate of the probability that the personality sketch was about Adam, if Adam was enrolled in a different field than the one in question:
\vspace{5pt}

agricultural and veterinary science
\vspace{5pt}

Please answer the question in the following format:
\vspace{5pt}

\{
  "agricultural and veterinary science": "your answer, please provide a value between 0 and 1"
\}
\end{promptbox}
\FloatBarrier

\begin{promptbox}[Inverse Probability for Alternative $P(E|\neg H)$ --- Rotation ABC]
\linespread{0.5}\selectfont
The following is a personality sketch.
\vspace{5pt}

The subject has a genuine curiosity about nature and its various plants. A quiet boy who enjoys solving puzzles, he has a warm heart, strong empathy, and a deep love for animals. He is of high intelligence and is good at understanding abstract concepts. The subject has a strong understanding and appreciation for the hard work and dedication.
\vspace{5pt}

In 2021, Adam was a graduate student in the U.S.
\vspace{5pt}

For each of the following fields, please provide your best estimate of the probability that the personality sketch was about Adam, if Adam was enrolled in a different field than the one in question:
\vspace{5pt}

business administration

agricultural and veterinary science

computer science
\vspace{5pt}

Your total does not need to add up to 1. Please answer the question in the following format:
\vspace{5pt}

\{
  "business administration": "your answer, please provide a value between 0 and 1",

  "agricultural and veterinary science": "your answer, please provide a value between 0 and 1",

  "computer science": "your answer, please provide a value between 0 and 1"
\}

\end{promptbox}
\FloatBarrier

\begin{promptbox}[Similarity --- Rotation A]
\linespread{0.5}\selectfont
The following is a personality sketch of Adam.
\vspace{5pt}

The subject has a genuine curiosity about nature and its various plants. A quiet boy who enjoys solving puzzles, he has a warm heart, strong empathy, and a deep love for animals. He is of high intelligence and is good at understanding abstract concepts. The subject has a strong understanding and appreciation for the hard work and dedication.
\vspace{5pt}

In 2021, Adam was a graduate student in the U.S.
\vspace{5pt}

For the following field, on a scale from 0 to 1, please indicate how similar you believe Adam is to a typical graduate student in that field, with 0 being the least similar and 1 being the most similar.
\vspace{5pt}

agricultural and veterinary science
\vspace{5pt}

Please answer the question in the following format:
\vspace{5pt}

\{
  "agricultural and veterinary science": "your answer, please provide a value between 0 and 1"
\}
\end{promptbox}
\FloatBarrier

\begin{promptbox}[Similarity --- Rotation ABC]
\linespread{0.5}\selectfont
The following is a personality sketch of Adam.
\vspace{5pt}

The subject has a genuine curiosity about nature and its various plants. A quiet boy who enjoys solving puzzles, he has a warm heart, strong empathy, and a deep love for animals. He is of high intelligence and is good at understanding abstract concepts. The subject has a strong understanding and appreciation for the hard work and dedication.
\vspace{5pt}

In 2021, Adam was a graduate student in the U.S.
\vspace{5pt}

For each of the following fields, on a scale from 0 to 1, please indicate how similar you believe Adam is to a typical graduate student in that field, with 0 being the least similar and 1 being the most similar.
\vspace{5pt}

business administration

computer science

agricultural and veterinary science
\vspace{5pt}

Please answer the question in the following format:
\vspace{5pt}

\{
  "business administration": "your answer, please provide a value between 0 and 1",

  "computer science": "your answer, please provide a value between 0 and 1",

  "agricultural and veterinary science": "your answer, please provide a value between 0 and 1"
\}
\end{promptbox}
\FloatBarrier

\subsection{Prompt Modifications in Unstructured Test}

In the prompt engineering for unstructured test, we changed the question for posterior judgment to:

\begin{enumerate}
\itemsep0em
\item For each of the following fields, please provide your best estimates of the posterior conditional probability that Adam was enrolled in that field given his personality sketch
\item For each of the following fields, please compute the posterior conditional probability that Adam was enrolled in that field given his personality sketch
\item Let E denote the personality sketch of Adam. For each of the following fields, denote H to be the hypothesis that Adam was enrolled in that field. Then, please compute the posterior conditional probability $P(H\mid E)$ using Bayes' rule. 
\item Let E denote the personality sketch of Adam. For each of the following fields, denote H to be the hypothesis that Adam was enrolled in that field. Now, please firstly judge P(H), $P(E\mid H)$ and $P(E\mid \text{\textbackslash neg}\ H)$, and then use them to compute $P(H\mid E)$ through Bayes rule.
\end{enumerate}

\section{Questioning Rounds Determination}

We model an LLM's response to a question as a random vector $\mathbf{X} = (X_{1}, \dots, X_{p})$. In semi-structured and unstructured tests, $\mathbf{X}$ is a continuous outcome, such as the posterior probability of each field. In structured tests, $X$ is an indicator function where $X\equiv 1$ if the answer is correct and $X\equiv 0$ if the answer is incorrect. 

What we are interested in is the population mean $\mathbb{E}(\mathbf{X})\equiv (\mu_{1},\dots, \mu_{p})$. Given a random sample $\mathbf{X}_{1}, \dots, \mathbf{X}_{n}$, the sample mean vector $\overline{\mathbf{X}} \equiv (\overline{x}_{1}, \dots, \overline{x}_{n})$, and the sample covariance matrix $\mathbf{S}$ with variances $s_{11}, \dots, s_{nn}$, we can construct the $100(1-\alpha)\%$ simultaneous confidence intervals for the population means provided $n - p$ is sufficiently large. These intervals are given by: $$\mu_{i} \in \overline{x}_{i} \pm \sqrt{\chi_{p}^{2}(\alpha)\frac{s_{ii}}{n}},$$ where $\chi_{p}^{2}(\alpha)$ denotes the chi-square critical value for $p$ degrees of freedom at a significance level $\alpha$ \cite[see, e.g.][\S 4.4]{cochran1977sampling}. In other words, given a predefined distance $d$ and significance level $\alpha$, our goal is to select an appropriate questioning round $n$ such that the following inequality is satisfied for all $i \in \{1, \dots, p\}$:
$$2\sqrt{\chi_{p}^{2}(\alpha)\frac{s_{ii}}{n}} \leq d.$$ 

To achieve this goal, a preliminary pilot study with a predetermined questioning round $n_{0}$ is initially conducted. Then, for each field $i \in \{1,\dots, p\}$, we could estimate the sample variance $s_{ii}$ and identify the field $k$ with the highest variance $s_{kk}$. Subsequently, we computed the following threshold value:
$$n_{1}:=\dfrac{4\chi_{p}^{2}(\alpha)s_{kk}}{d^{2}}.$$ 
If $n_{0} \geq n_{1}$, then $n_{0}$ is deemed sufficient. Conversely, if $n_{0} < n_{1}$, then a further pilot study should be conducted with the questioning round $n_{1}$.

It is noteworthy that this approach does not require the assumption of a normal population as our random samples are independent and identically distributed (i.i.d.). In our experimental setup, the LLM was reset with no prior memory before each round of questioning, ensuring that interactions in previous rounds do not influence responses in subsequent rounds.\footnote{Communication with the LLMs via API is guaranteed not to be used for training purposes by the provider. Even if it had been used, the experimental duration was too brief to allow for any training substantial enough to impact our experimental results.} Furthermore, for each model, every response to the questioning is derived from the same distribution induced by the same LLM and the same question, thereby ensuring that the samples are identically distributed.

Our experiments encompass four types of responses: an accuracy rate (the mean of the indicator random variable) within $[0,1]$, a similarity score within $[0,1]$, a probability within $[0,1]$, and a percentage estimate within $[0,100]$. For each of these metrics, we have fixed $\alpha=0.01$, ensuring a $99\%$ confidence level. The distance parameter is set to $d=0.05$ for the first three metrics and $d=3$ for the percentage estimate. This setup guarantees that the true mean lies within $\pm 0.025$ of the sample mean for the first three metrics and within $\pm 1.5$ for the percentage estimate. The selection of these parameters reflects a balance between the statistical robustness of our results and our available resources. Furthermore, the initial pilot study requires a sample size $n_{0}$ sufficiently large to capture enough variability. For instance, if $n_{0}=5$ and one obtains fortuitously low variance, it may falsely suggest adequacy of $n_{0}$. Consequently, in our experiment, we set $n_{0}=100$ for the structured test, $300$ for the semi-structured test, and $150$ for the unstructured test.

\subsection{Structured Test}

Now, we provide details of sample size for each model in each question. For structured test, again a blank space indicates that the model either declined to provide a definitive answer or failed to deliver the response in a consistent and parsable format.

\begin{table}[htbp]
\centering
\caption{Structured Test: Questioning Rounds}
\resizebox{\columnwidth}{!}{%
\begin{tabular}{lllllllllllllllll}
\hline
 &
  \textbf{Q1} &
  \textbf{Q2} &
  \textbf{Q3} &
  \textbf{Q4} &
  \textbf{Q5} &
  \textbf{Q6} &
  \textbf{Q7} &
  \textbf{Q8} &
  \textbf{Q9} &
  \textbf{Q10} &
  \textbf{Q11} &
  \textbf{Q12} &
  \textbf{Q13} &
  \textbf{Q14} &
  \textbf{Q15} &
  \textbf{Q16} \\ \hline
gpt-4o                   & 100 & 152 & 120 & 100 & 100 & 100 & 150 & 330 & 300 & 300 & 150  & 150  & 150 & 150 & 150 & 150 \\
gpt-4-turbo              & 100 & 152 & 120 & 100 & 100 & 100 & 150 & 370 & 300 & 300 & 150  & 150  & 150 & 150 & 150 & 150 \\
gpt-3.5-turbo-0125       & 100 & 152 & 120 & 100 & 139 & 100 & 520 & 370 & 300 & 300 & 280  & 380  & 150 & 150 & 190 & 150 \\
gemini-1.5-pro-latest    & 100 & 152 & 120 & 100 & 100 & 100 & 290 & 150 & 300 & 300 &      & 150  & 150 & 150 & 150 & 150 \\
gemini-1.5-flash-latest  & 100 & 152 & 120 & 100 & 100 & 100 & 150 & 300 & 300 & 300 & 150  & 150  & 150 & 150 & 150 & 150 \\
gemini-1.0-pro-latest    & 105 & 152 & 120 & 100 & 100 & 100 & 500 & 300 & 300 & 300 & 150  & 570  & 150 & 150 & 150 & 150 \\
claude-3-opus-20240229   & 100 & 152 & 120 & 100 & 100 & 100 & 150 & 430 & 300 & 300 & 150  & 150  & 150 & 150 & 150 & 150 \\
claude-3-sonnet-20240229 & 100 & 152 & 120 & 100 & 100 & 100 & 150 & 150 & 300 & 330 & 150  & 150  & 150 & 150 & 150 & 150 \\
claude-3-haiku-20240307  & 100 & 152 & 120 & 100 & 199 & 100 & 270 & 270 & 300 & 300 & 150  & 150  & 150 & 150 & 150 & 150 \\
open-mixtral-8x22b       &     & 152 & 120 & 100 & 100 & 100 & 150 & 150 & 300 & 300 & 150  & 150  & 150 & 150 & 150 & 150 \\
open-mixtral-8x7b        & 100 & 152 & 120 & 100 & 100 & 100 & 600 & 600 & 300 & 680 & 350  & 150  & 150 & 150 & 150 & 150 \\
open-mistral-7b          & 100 & 152 & 120 &     &     &     &     &     &     &     & 1000 & 1000 &     &     &     &     \\
mistral-large-latest     & 100 & 152 & 120 & 100 & 100 & 100 & 150 & 480 & 300 & 300 & 150  & 150  & 150 & 150 & 150 & 150 \\
mistral-medium-latest    & 100 & 152 & 120 & 100 & 100 & 100 & 150 & 570 & 300 & 300 & 150  & 150  &     &     &     &     \\
mistral-small-latest     &     & 152 & 120 & 100 & 100 & 100 & 480 & 200 & 300 & 300 & 170  & 240  & 150 & 150 & 150 & 150 \\ \hline
\end{tabular}%
}
\end{table}
\FloatBarrier

\subsection{Semi-Structured Test}

\

\begin{longtable}[c]{lcccccc}
\caption{Semi-Structured Test: Questioning Rounds}\\
\hline\hline
 &
  \multicolumn{2}{c}{Uncharacteristic} &
  \multicolumn{2}{c}{\begin{tabular}[c]{@{}c@{}}Representative \\ Computer Science\end{tabular}} &
  \multicolumn{2}{c}{\begin{tabular}[c]{@{}c@{}}Representative \\ Humanities\end{tabular}} \\ \cline{2-7} 
\endfirsthead
\endhead
\hline
\endfoot
\endlastfoot
\textbf{} &
  \begin{tabular}[c]{@{}c@{}}High \\ Base Rate\end{tabular} &
  \begin{tabular}[c]{@{}c@{}}Low \\ Base Rate\end{tabular} &
  \begin{tabular}[c]{@{}c@{}}High \\ Base Rate\end{tabular} &
  \begin{tabular}[c]{@{}c@{}}Low \\ Base Rate\end{tabular} &
  \begin{tabular}[c]{@{}c@{}}High \\ Base Rate\end{tabular} &
  \begin{tabular}[c]{@{}c@{}}Low \\ Base Rate\end{tabular} \\ \cline{2-7} 
gpt-4o             & 300 & 300 & 300 & 300 & 300 & 300 \\
gpt-4-turbo        & 300 & 300 & 300 & 300 & 300 & 300 \\
gpt-3.5-turbo      & 300 & 300 & 300 & 300 & 300 & 300 \\
gemini-1.5-pro     & 300 & 300 & 300 & 300 & 300 & 300 \\
gemini-1.5-flash   & 300 & 300 & 300 & 300 & 300 & 300 \\
gemini-1.0-pro     & 300 & 300 & 300 & 300 & 300 & 300 \\
claude-3-opus      & 300 & 300 & 300 & 300 & 300 & 300 \\
claude-3-haiku     & 300 & 300 & 300 & 300 & 300 & 300 \\
claude-3-sonnet    & 300 & 330 & 300 & 550 & 300 & 300 \\
open-mixtral-8x22b & 300 & 300 & 300 & 300 & 300 & 300 \\
open-mixtral-8x7b  & 300 & 300 & 300 & 300 & 300 & 300 \\
open-mistral-7b    & 300 & 300 & 300 & 300 & 300 & 300 \\
mistral-large      & 300 & 300 & 300 & 300 & 300 & 300 \\
mistral-medium     & 300 & 300 & 300 & 300 & 300 & 300 \\
mistral-small      & 300 & 300 & 300 & 300 & 300 & 300 \\ \hline\hline
\end{longtable}
\FloatBarrier

\begin{table}[htbp]
\centering
\caption{Semi-Structured Test Prompt Engineering: Questioning Rounds of GPT-4o}
\resizebox{\columnwidth}{!}{%
\begin{tabular}{lcccccc}
\hline
 &
  \multicolumn{2}{c}{Uncharacteristic} &
  \multicolumn{2}{c}{\begin{tabular}[c]{@{}c@{}}Representative \\ Computer Science\end{tabular}} &
  \multicolumn{2}{c}{\begin{tabular}[c]{@{}c@{}}Representative \\ Humanities\end{tabular}} \\ \cline{2-7} 
 &
  \begin{tabular}[c]{@{}c@{}}High \\ Base Rate\end{tabular} &
  \begin{tabular}[c]{@{}c@{}}Low \\ Base Rate\end{tabular} &
  \begin{tabular}[c]{@{}c@{}}High \\ Base Rate\end{tabular} &
  \begin{tabular}[c]{@{}c@{}}Low \\ Base Rate\end{tabular} &
  \begin{tabular}[c]{@{}c@{}}High \\ Base Rate\end{tabular} &
  \begin{tabular}[c]{@{}c@{}}Low \\ Base Rate\end{tabular} \\ \cline{2-7} 
Modification 1 & 150 & 150 & 150 & 150 & 210 & 150 \\
Modification 2 & 150 & 150 & 150 & 150 & 150 & 150 \\
Modification 3 & 150 & 150 & 150 & 150 & 150 & 150 \\ \hline
\end{tabular}%
}
\end{table}
\FloatBarrier

\subsection{Unstructured Test}

\

\begin{longtable}[c]{llllllll}
\caption{Unstructured Test: Questioning Rounds}\\
\hline
 &
  \multicolumn{1}{c}{\textbf{A}} &
  \multicolumn{1}{c}{\textbf{B}} &
  \multicolumn{1}{c}{\textbf{C}} &
  \multicolumn{1}{c}{\textbf{AB}} &
  \multicolumn{1}{c}{\textbf{AC}} &
  \multicolumn{1}{c}{\textbf{BC}} &
  \multicolumn{1}{c}{\textbf{ABC}} \\ \hline
\endfirsthead
\endhead
\hline
\endfoot
\endlastfoot
                                            & \multicolumn{7}{c}{\textbf{Prior}}                                 \\ \hline
\multicolumn{1}{l|}{gpt-4o}                 & 150      & 150      & 150     & 150     & 150    & 150    & 150    \\
\multicolumn{1}{l|}{gpt-4-turbo}            & 150      & 150      & 150     & 150     & 150    & 150    & 150    \\
\multicolumn{1}{l|}{gemini-1.5-pro-latest}  & 150      & 150      & 150     & 150     & 150    & 150    & 150    \\
\multicolumn{1}{l|}{claude-3-opus-20240229} & 150      & 150      & 150     & 150     & 150    & 150    & 150    \\
\multicolumn{1}{l|}{mistral-large-latest}   & 150      & 150      & 150     & 150     & 150    & 150    & 150    \\ \hline
                                            & \multicolumn{7}{c}{\textbf{Inverse Probability}}                   \\ \hline
\multicolumn{1}{l|}{gpt-4o}                 & 150      & 150      & 230     & 150     & 200    & 150    & 350    \\
\multicolumn{1}{l|}{gpt-4-turbo}            & 150      & 150      & 150     & 150     & 190    & 150    & 240    \\
\multicolumn{1}{l|}{gemini-1.5-pro-latest}  & 150      & 150      & 150     & 150     & 150    & 150    & 150    \\
\multicolumn{1}{l|}{claude-3-opus-20240229} & 150      & 150      & 150     & 150     & 150    & 150    & 150    \\
\multicolumn{1}{l|}{mistral-large-latest}   & 150      & 150      & 150     & 150     & 150    & 150    & 170    \\ \hline
                                            & \multicolumn{7}{c}{\textbf{Inverse Probabiliy of The Alternative}} \\ \hline
\multicolumn{1}{l|}{gpt-4o}                 & 180      & 150      & 150     & 150     & 150    & 510    & 360    \\
\multicolumn{1}{l|}{gpt-4-turbo}            & 150      & 150      & 150     & 150     & 150    & 150    & 760    \\
\multicolumn{1}{l|}{gemini-1.5-pro-latest}  & 150      & 150      & 150     & 150     & 150    & 390    & 150    \\
\multicolumn{1}{l|}{claude-3-opus-20240229} & 150      & 150      & 230     & 150     & 150    & 150    & 200    \\
\multicolumn{1}{l|}{mistral-large-latest}   & 150      & 150      & 150     & 150     & 150    & 150    & 150    \\ \hline
                                            & \multicolumn{7}{c}{\textbf{Posterior}}                             \\ \hline
\multicolumn{1}{l|}{gpt-4o}                 & 260      & 150      & 150     & 150     & 150    & 150    & 150    \\
\multicolumn{1}{l|}{gpt-4-turbo}            & 150      & 150      & 150     & 150     & 150    & 220    & 260    \\
\multicolumn{1}{l|}{gemini-1.5-pro-latest}  & 150      & 150      & 150     & 150     & 150    & 278    & 150    \\
\multicolumn{1}{l|}{claude-3-opus-20240229} & 150      & 150      & 200     & 190     & 150    & 380    & 260    \\
\multicolumn{1}{l|}{mistral-large-latest}   & 150      & 150      & 150     & 150     & 150    & 150    & 190    \\ \hline
                                            & \multicolumn{7}{c}{\textbf{Similarity}}                            \\ \hline
\multicolumn{1}{l|}{gpt-4o}                 & 150      & 150      & 150     & 150     & 150    & 150    & 150    \\
\multicolumn{1}{l|}{gpt-4-turbo}            & 150      & 150      & 150     & 150     & 150    & 150    & 150    \\
\multicolumn{1}{l|}{gemini-1.5-pro-latest}  & 150      & 150      & 150     & 150     & 150    & 390    & 150    \\
\multicolumn{1}{l|}{claude-3-opus-20240229} & 150      & 150      & 150     & 150     & 150    & 150    & 150    \\
\multicolumn{1}{l|}{mistral-large-latest}   & 150      & 150      & 150     & 150     & 150    & 150    & 150    \\ \hline
\end{longtable}
\FloatBarrier

\begin{longtable}[c]{llllllll}
\caption{Unstructured Test Prompt Engineering: Questioning Rounds of GPT-4o}
\\
\hline
               & \textbf{A} & \textbf{B} & \textbf{C} & \textbf{AB} & \textbf{AC} & \textbf{BC} & \textbf{ABC} \\ \hline
\endfirsthead
\endhead
\hline
\endfoot
\endlastfoot
Modification 1 & 150        & 150        & 150        & 150         & 150         & 290         & 150          \\
Modification 2 & 260        & 150        & 320        & 150         & 190         & 300         & 200          \\
Modification 3 & 590        & 150        & 370        & 670         & 650         & 560         & 470          \\
Modification 4 & 270        & 150        & 150        & 1360        & 680         & 1170        & 950          \\ \hline
\end{longtable}
\FloatBarrier

\end{document}